\documentclass[conference]{IEEEtran}
\IEEEoverridecommandlockouts
\usepackage{cite}
\usepackage{amsmath,amssymb,amsfonts}
\usepackage{algorithmic}
\usepackage{graphicx}
\usepackage{textcomp}
\usepackage{xcolor}
\usepackage{url}

\def\BibTeX{{\rm B\kern-.05em{\sc i\kern-.025em b}\kern-.08em
    T\kern-.1667em\lower.7ex\hbox{E}\kern-.125emX}}
\begin{document}

\title{Applying Machine Learning Tools for Urban Resilience Against Floods}

\author{
\IEEEauthorblockN{Mahla Ardebili Pour\textsuperscript{*}}
\IEEEauthorblockA{\textit{Civil and Environmental Engineering} \\
\textit{University of California, Davis}\\
Davis, USA \\
Mardebilipour@ucdavis.edu}
\and
\IEEEauthorblockN{Mohammad B. Ghiasi}
\IEEEauthorblockA{\textit{Electrical and Computer Engineering} \\
\textit{University of California, Davis}\\
Davis, USA \\
Mbghiasi@ucdavis.edu}
\and
\IEEEauthorblockN{Ali Karkehabadi}
\IEEEauthorblockA{\textit{Electrical and Computer Engineering} \\
\textit{University of California, Davis}\\
Davis, USA \\
Akarkehabadi@ucdavis.edu}
}

\maketitle

\begingroup
\renewcommand\thefootnote{\textsuperscript{*}}
\footnotetext{Corresponding author: Mahla Ardebili Pour (Mardebilipour@ucdavis.edu)}
\endgroup

\begin{abstract}
Floods are among the most prevalent and destructive natural disasters, often leading to severe social and economic impacts in urban areas due to the high concentration of assets and population density. In Iran, particularly in Tehran, recurring flood events underscore the urgent need for robust urban resilience strategies. This paper explores flood resilience models to identify the most effective approach for District 6 in Tehran. Through an extensive literature review, various resilience models were analyzed, with the Climate Disaster Resilience Index (CDRI) emerging as the most suitable model for this district due to its comprehensive resilience dimensions: Physical, Social, Economic, Organizational, and Natural/Health resilience. Although the CDRI model provides a structured approach to resilience measurement, it remains a static model focused on spatial characteristics and lacks temporal adaptability. An extensive literature review enhances the CDRI model by integrating data from 2013 to 2022 in three-year intervals and applying machine learning techniques to predict resilience dimensions for 2025. This integration enables a dynamic resilience model that can accommodate temporal changes, providing a more adaptable and data-driven foundation for urban flood resilience planning. By employing artificial intelligence to reflect evolving urban conditions, this model offers valuable insights for policymakers and urban planners to enhance flood resilience in Tehran’s critical District 6.

\end{abstract}

\begin{IEEEkeywords}
Resiliency, CDRI, Flood, Urban Resilience, Machine Learning
\end{IEEEkeywords}

\section{Introduction}
In March 2019, catastrophic floods in Iran led to the loss of over 70 lives and resulted in around 2.5\$ billion in damages. Similarly, historical flooding events in Tehran from 1987 to 2015 highlight the city's vulnerability to such disasters \footnote{The International Disaster Database. \url{https://www.emdat.be/}}. To mitigate the impact of future floods, enhancing Tehran’s resilience is essential. This resilience plan is especially critical for areas like District 6, recognized as Tehran’s "beating heart" and a central hub of activity, population growth, and urban development \cite{b3}. Crises—whether natural or human-made—disrupt societal systems, often isolating individuals and communities from planned responses \cite{b4}. These unforeseen events generally have severe effects on people, the environment, and the economy. The growing frequency and variety of natural disasters represent a significant challenge to sustainable development. Vulnerability to these events affects much of the world, leading to physical destruction, social instability, and financial losses. Addressing these vulnerabilities requires focused efforts to strengthen resilience in impacted communities \cite{b4}. Developing a framework to enhance urban resilience, especially in flood-prone areas, has become essential, as studies suggest \cite{norouzi2021framework, liang2020resilient, ma2021resilience}. Climate change, though a global issue, has particularly severe effects in certain regions, including Iran, where it contributes to frequent droughts and sudden floods that annually inflict significant harm on both natural and human resources \cite{rahimi2023climate}. Since the 1980s, climate-related disasters such as floods and hurricanes have surged worldwide, underscoring the growing importance of urban resilience. While natural factors undeniably play a role in flood occurrences, human activities (such as environmental degradation and resource depletion) also exacerbate the intensity of these disasters, increasing the potential for destruction.

Enhancing resilience is a critical step toward achieving sustainability, particularly in a country prone to severe natural hazards like floods and earthquakes, which have historically caused significant financial damage and considerable loss of life due to environmental and geographic vulnerabilities. Resilience is defined as "the capacity of a complex system to recover following a natural or human-made disaster" \cite{b4}. This paper represents an initial exploration of resilience strategies. One effective approach to mitigating urban flood risks involves capturing urban runoff and repurposing it for non-potable uses. As urbanization expands, the risks of flooding increase, creating an urgent need for both flood management and green space development. Utilizing urban runoff as a supplemental water source for irrigation in green spaces addresses these needs by reducing flood risks and supporting urban sustainability efforts.

\section{Related Work}
Floods are defined by the U.S. Environmental Protection Agency (EPA) as the temporary or partial inundation of normally dry land, caused by factors like heavy rainfall, inland or tidal water overflow, or rapid surface runoff \cite{b6, jones2020flood}. Similarly, the European Union’s Flood Directive and Chow’s work describe floods as the overflow of water breaching natural or artificial barriers. The U.S. Geological Survey (USGS) categorizes floods into river floods, which mainly cause property damage, and flash floods, which pose a greater risk to human life due to their sudden onset \cite{b4}. In Iran, floods may result from intense rainfall, combined rain and snowmelt, or isolated snowmelt, and are classified into sea wave floods, flash floods, and river floods \cite{b4, smith2021classification, taylor2022flood}. Globally, floods are among the most destructive natural hazards, causing significant loss of life and economic damage. Between 1988 and 1997, floods accounted for 58\% of disaster-related fatalities worldwide, with major historical floods like the 1931 Yellow River flood in China resulting in casualties between 800,000 and 4 million \cite{b9, b9-1}. The concept of resilience, from the Latin \textit{resilio} (“to spring back”), is critical in urban studies and disaster management, referring to the ability of a system to absorb shocks while maintaining core functions \cite{b11}. In urban settings, resilience encompasses the capacity of individuals, communities, and ecosystems to withstand, recover from, and adapt to stresses \cite{b12}. The United Nations International Strategy for Disaster Reduction (UNISDR) emphasizes resilience as essential for effective disaster response and recovery, balancing disaster preparedness with vulnerability reduction \cite{b13, adger2021social}. Several quantitative models of resilience have been developed to improve urban adaptability to climate challenges. Shaw et al. introduced the Climate Disaster Resilience Index (CDRI), which evaluates urban resilience across five dimensions and has been applied in several Asian countries, including Japan and India \cite{b14}. Tyler and Munch's Urban Resilience Framework (URF) views resilience as a system's capacity to maintain performance under stress \cite{b15}. Cutter et al. developed the Baseline Resilience Indicators for Communities (BRIC) for the southeastern U.S. \cite{b16}. Other models include Renschler's Community Resilience Framework (PEOPLES), the Coastal Community Adaptation Resilience (CCaR) model for coastal flood-prone areas \cite{b17, b18}, and the Community Climate Resilience Assessment Measure (CCRAM), tested in diverse Israeli communities \cite{b19}.

\begin{table}[h!]
\centering
\caption{Comparison of Resilience Models}
\vspace{-8pt}
\label{Table}
\renewcommand{\arraystretch}{1.2}
\resizebox{0.95\columnwidth}{!}{%
\begin{tabular}{|p{2cm}|p{4cm}|p{2.5cm}|}
\hline
\textbf{Model Name} & \textbf{Indicators} & \textbf{Scale} \\ \hline
CDRI\cite{joerin2011chapter} & Physical, Social, Economic, Organizational, Natural & City/Region \\ \hline
URF \cite{b15} & Urban systems (ecosystem, infrastructure, organization) & City/District \\ \hline
PEOPLES\cite{b17} & Population, Environment, Gov. Services, Infrastructure, Lifestyle, Economy, Social & Multi-City \\ \hline
BRIC\cite{b16} & Social, Economic, Institutional, Infrastructure, Social Capital & County \\ \hline
STDRM\cite{b18} & Economic, Health, Organizational, Physical, Social & County \\ \hline
CCRAM\cite{b19} & Leadership, Collective Efficacy, Preparedness, Trust, Relationship & Societies \\ \hline
SERV\cite{b20} & Sensitivity, Exposure, Adaptive Capacity & County \\ \hline
WISC\cite{b21} & Welfare, Identity, Social Services, Social Capitals & Communities \\ \hline
REDI\cite{b22} & Physical, Social, Natural & Local/Regional \\ \hline
S-DReP\cite{b23} & Spatio-environmental & High-density Cities \\ \hline
\end{tabular}%
}
\end{table}

Resilience models tailored for specific contexts highlight the importance of addressing unique regional needs. Frazier’s Social Vulnerability and Economic Resilience (SERV) model, developed for Florida, emphasizes reducing vulnerability and enhancing economic resilience at urban and regional scales \cite{b20}. The WISC model by Miles, with a focus on well-being and community cohesion, includes 29 indicators to assess local resilience \cite{b21}. Kontokosta’s Resilience to Disasters Index (REDI), validated with Hurricane Sandy data, provides insights into resilience for disaster-prone areas \cite{b22}. In urban flood management, resilience is bolstered by controlling surface runoff. Key criteria for underground runoff reservoir locations were identified through expert feedback and structured questionnaires, encompassing environmental factors (e.g., proximity to waste centers and green spaces), topographic features (e.g., fault distance and slope), and land use considerations (e.g., proximity to highways and metro stations). Variables like soil type and geomorphology, though important, were excluded due to data limitations. Analyzing historical flood-prone areas supports proactive urban planning and strengthens resilience against future flood risks \cite{godschalk2003, miller2022community}.

 Machine learning (ML) has emerged as a useful tool in comprehensive areas, such as power efficiency and noise reduction\cite{ghiasi2022simple,shei}, and resilience modelling, especially for analyzing complex, large-scale datasets \cite{karkehabadi2024smoot, karkehabadihlgm, bakhshi2024novel, hassanpour2024overcoming}. ML In this research, ML techniques are applied to enhance the temporal dynamics of the Climate Disaster Resilience Index (CDRI). ML's predictive capabilities, using historical data, enable urban resilience assessments that inform proactive urban planning and policy-making for climate-related challenges. To project urban resilience indicators for 2025, historical data from 2013 to 2022 are used at three-year intervals. This approach builds on previous CDRI-based research, adapting the index to evolving conditions. Several ML models are utilized, including:

\begin{itemize}
    \item \textbf{Linear Regression} serves as a baseline for trend observation but is limited to modeling linear relationships \cite{montgomery2014applied}.
    \item \textbf{Decision Trees} highlight feature importance, offering interpretability in resilience predictions while necessitating pruning to mitigate overfitting risks \cite{breiman1984classification}.
    
    \item \textbf{Random Forests}, an ensemble method, enhance stability by averaging predictions, aiding complex datasets \cite{ho1995random}.
    \item \textbf{Gradient Boosting} improves accuracy iteratively by targeting underperforming areas, making it ideal for time-series analysis but requiring careful tuning to avoid overfitting \cite{friedman2001greedy, karkehabadi2024ffcl}.

    \item \textbf{Vector Autoregression (VAR)} models temporal interdependencies, enabling nuanced resilience analysis across indicators \cite{lutkepohl2005new}.
    \item \textbf{Long Short-Term Memory (LSTM)} networks are optimized for sequential data, capturing complex temporal dependencies for long-term resilience forecasting \cite{hochreiter1997long}.
\end{itemize}

Through these models, this study seeks to advance predictive modeling for urban resilience, providing data-driven insights that support urban planners and policymakers in strengthening climate resilience at the urban level.

\section{Materials and Methods}
To explore the concept of resilience and identify various resilience models, a thorough literature review was conducted. Studies in resilience encompass both qualitative and quantitative approaches, though most of the research predominantly uses qualitative methods, with relatively few studies focusing on quantitative analysis. Moreover, most of the review articles available in the Scopus database center around quality-focused analyses, which has led to a need for more quantitative research. Consequently, this study emphasizes quantitative methods to assess urban resilience. Various frameworks and key studies on resilience were examined and compared, focusing on identifying optimal locations for reservoirs. Geographic Information System (GIS) modeling was then applied to aid this task. Additionally, machine learning (ML) methods were integrated with the Climate Disaster Resilience Index (CDRI) to enhance resilience models for this region. By applying ML to the CDRI, the model becomes more realistic, factoring in temporal dynamics and adaptive responses over time. This approach allows the CDRI to account for evolving conditions and provides a more data-driven, precise foundation for planning reservoir site selection, ultimately improving resilience assessment and decision-making.

\begin{figure}[h!]
\centering
\includegraphics[width=0.4\textwidth]{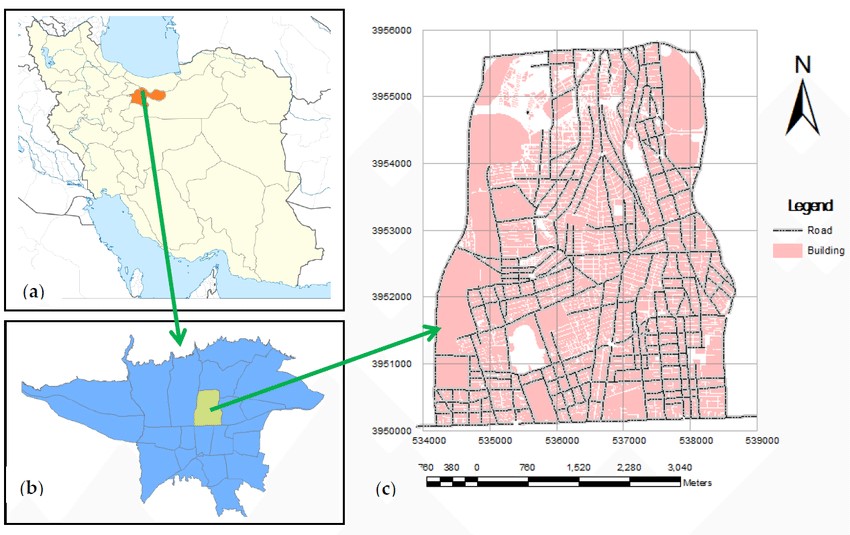}
\caption{a) Iran, b) Tehran, c) Region 6 of Tehran }
\label{fig:map}
\end{figure}
District 6 (i.e., Figure~\ref{fig:map})  holds a significant spatial position due to its rich historical heritage, including landmarks such as the Bazaar and Topkhaneh Square. This area has a population of 251,384 residents (based on data from the 2016 census) and covers approximately 2,137.9 hectares. Geographically, District 6 lies at a longitude of 51.40 a latitude of 35.7, and an altitude of around 1300 meters above sea level. The region is bordered by Hemmat Highway to the north, Azadi-Enghelab to the south, Modares and Shahid Mofteh highways to the east, and Shahid Chamran highway to the west. Although it makes up only 3\% of Tehran’s land area and 2.9\% of its population, District 6 contains over 30\% of the city’s governmental, public, and private buildings, including key national administrative offices, banks, and institutions. This concentration positions District 6 as the core of Tehran’s governance, public administration, and economic operations. The district is divided into six zones with 18 neighborhoods. The presence of student dormitories has significantly influenced the social and demographic profile of District 6, which is primarily characterized by administrative and commercial developments. Key landmarks include parks such as Laleh Park, Saei Park, and Ganjavi Military Park, as well as notable sites like Saint Maryam Church, the Museum of Contemporary Art, and major educational institutions like Amirkabir University of Technology and the University of Tehran \cite{b3}. Given its urban and socio-economic significance, it is crucial to prioritize the inspection and maintenance of these essential buildings and public spaces \cite{b3, farzanegan2020urban, mirzaei2018resilience}.

\section{Methodology}

This study presents a comparative analysis of key research on urban resilience to flooding from 2010 to 2021. A comprehensive literature review was conducted to identify methodologies applicable to urban flood resilience. Based on insights from this review and considering the specific characteristics and requirements of Tehran’s urban structure, the Climate Disaster Resilience Index (CDRI) was selected as the optimal approach for estimating resilience in this context. The CDRI method enables data collection through structured questionnaires, providing targeted resilience indicators tailored to the study area\cite{pour2024urban}. Additionally, machine learning (ML) techniques were incorporated into the CDRI model to extend resilience estimation over time. This integration allows the resilience model to capture temporal dynamics, leading to a more adaptive and realistic assessment of resilience.

Numerous studies have developed various resilience models. Eight main resilience models were evaluated based on criteria such as indicators, model structure, scale, and conceptual framework (Table~\ref{Table}). This review reveals that many existing frameworks remain theoretical, lacking validation with real-world data. A significant number of these models still require practical testing and refinement to ensure their effectiveness in real-world applications.

\subsection{Predicting Urban Resilience Features for 2025 by Using Machine Learning Models}
This study uses various machine learning models to predict urban resilience features for 2025, leveraging historical data from three-year intervals from 2013 to 2022. This work builds upon prior research on the CDRI method, enhancing it by making the CDRI dynamic over time.

\begin{table*}[h]
\centering
\caption{Presents 2025 resilience feature predictions generated by each machine learning model, facilitating a comparative performance analysis.}
\label{tab:prediction_results}
\begin{tabular}{|l|c|c|c|c|c|}
\hline
\textbf{Model} & \textbf{Physical} & \textbf{Social} & \textbf{Economic} & \textbf{Organizational} & \textbf{Natural/Health} \\
\hline
Linear Regression & 4.03092 & 4.47864& 2.02203 & 2.85259 & 2.75703 \\
Decision Tree & 4.04375 & 4.26504 &  2.16125 & 2.74750 & 2.65250 \\
Random Forest & 4.08358 & 4.24559 & 2.16961 & 2.75330  & 2.71039 \\
Gradient Boosting & 4.04375 & 4.26499 & 2.16125 & 2.78749 & 2.71249\\
VAR &  4.11138 & 4.39719 & 2.09061 & 2.59173 & 2.34924 \\
LSTM & 4.20363 & 4.50197 & 2.13022 & 2.79242 &  2.46654 \\
\hline
\end{tabular}
\end{table*}

\subsection{Data Collection and Structure}
The study involved structured questionnaires with 11 experts from relevant organizations, gathering insights from three-year intervals from 2013 to 2022. The assessment for Tehran’s District 6 utilized a questionnaire centered on five resilience dimensions—physical, social, economic, institutional, and natural—each with five parameters rated on a scale from 1 (weak) to 5 (good). Parameters were also weighted by importance to ensure alignment with local district conditions. The parameters in each dimension were ranked and weighted to prioritize variables specific to District 6's needs. This approach produced 44 spider diagrams that summarize annual resilience assessments by each expert across four years, allowing for a comparative analysis of resilience trends over time. These diagrams visually represent resilience scores for each dimension, highlighting temporal shifts and resilience strengths across assessment years.
The data used in this study was gathered from official government agencies involved in urban infrastructure management, including departments responsible for water, health, sewage, and transportation systems. This dataset provides a comprehensive view of urban resilience across five dimensions—physical, social, economic, organizational, and natural/health—captured at three-year intervals from 2013 to 2022.

\begin{figure}[h!]
\centering
\includegraphics[width=0.36\textwidth]{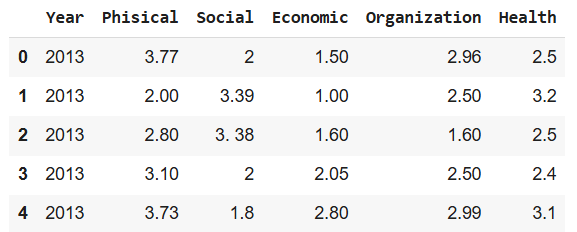}
\vspace{-5pt}

\caption{Illustrates scores across multiple dimensions over the assessment 2013 year}
\label{fig:data_sample}
\end{figure}

Figure~\ref{fig:data_sample} provides a sample of this dataset, illustrating the resilience scores for each feature across the 2013 year.

\section{Results}
Resilience is a developing concept, and no standardized framework yet exists for assessing resilience against natural disasters. Initial models suggest that reducing disaster risks and vulnerabilities can enhance community resilience by strengthening local capacities and adopting innovative strategies. Most resilience models focus on similar factors, including skills, resources, social networks, economic stability, and access to essential services, all of which contribute to reducing vulnerabilities and building community resilience. Social capital is central to these models, as it connects community wealth to resilience. Among the various models, the Spatio-temporal Disaster Resilience Model (STDRM) offers a robust temporal-spatial approach, while the PEOPLES model provides a comprehensive range of indicators. The CDRI model stands out for its extensive datasets, emphasizing its suitability for assessing urban resilience in diverse contexts.

In 2019, thel CDRI model captured a decline in the economic and helath resilience of District 6 in Tehran, reflecting the effects of COVID-19 and inflation and economic challenges in Iran that year. This dimension-specific analysis illustrates the model’s capability to track resilience changes over time, providing deeper insights into the effects of external socio-economic pressures on urban resilience. The effectiveness of the dynamic, AI-integrated CDRI model emphasizes the importance of adaptable, data-driven assessments for targeted policy interventions, enhancing resilience planning for flood-prone urban areas like Tehran.

\subsubsection{Prediction Results for 2022 based on Machine Learning}
\begin{figure}[h!]
\centering
\includegraphics[width=0.37\textwidth]{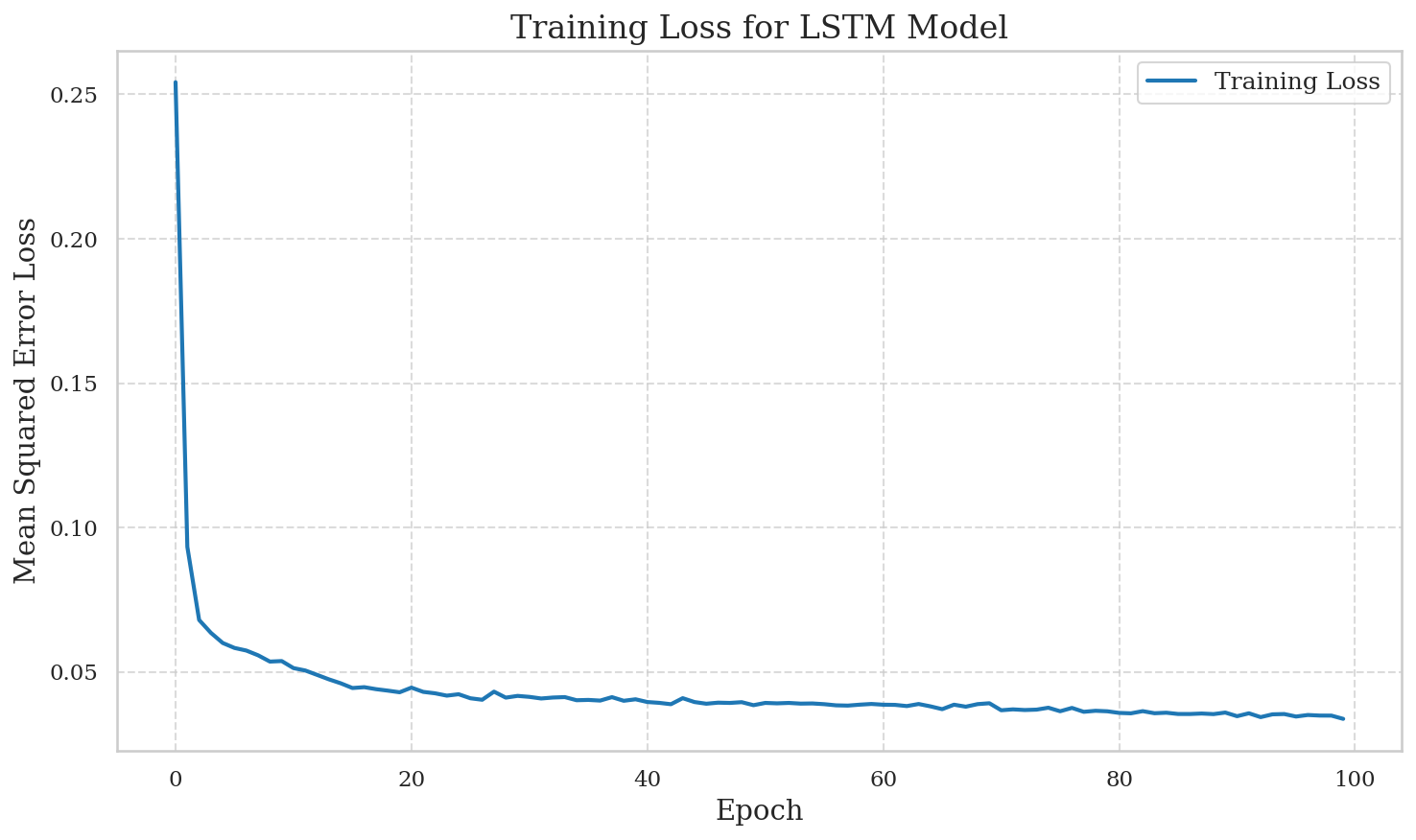}
\vspace{-5pt}
\caption{Depicts the reduction in loss during the LSTM model's training process, demonstrating how accuracy improves over time.}
\label{fig:lstm_loss}
\end{figure}

\begin{figure*}[h!]
\centering
\includegraphics[width=0.805\textwidth]{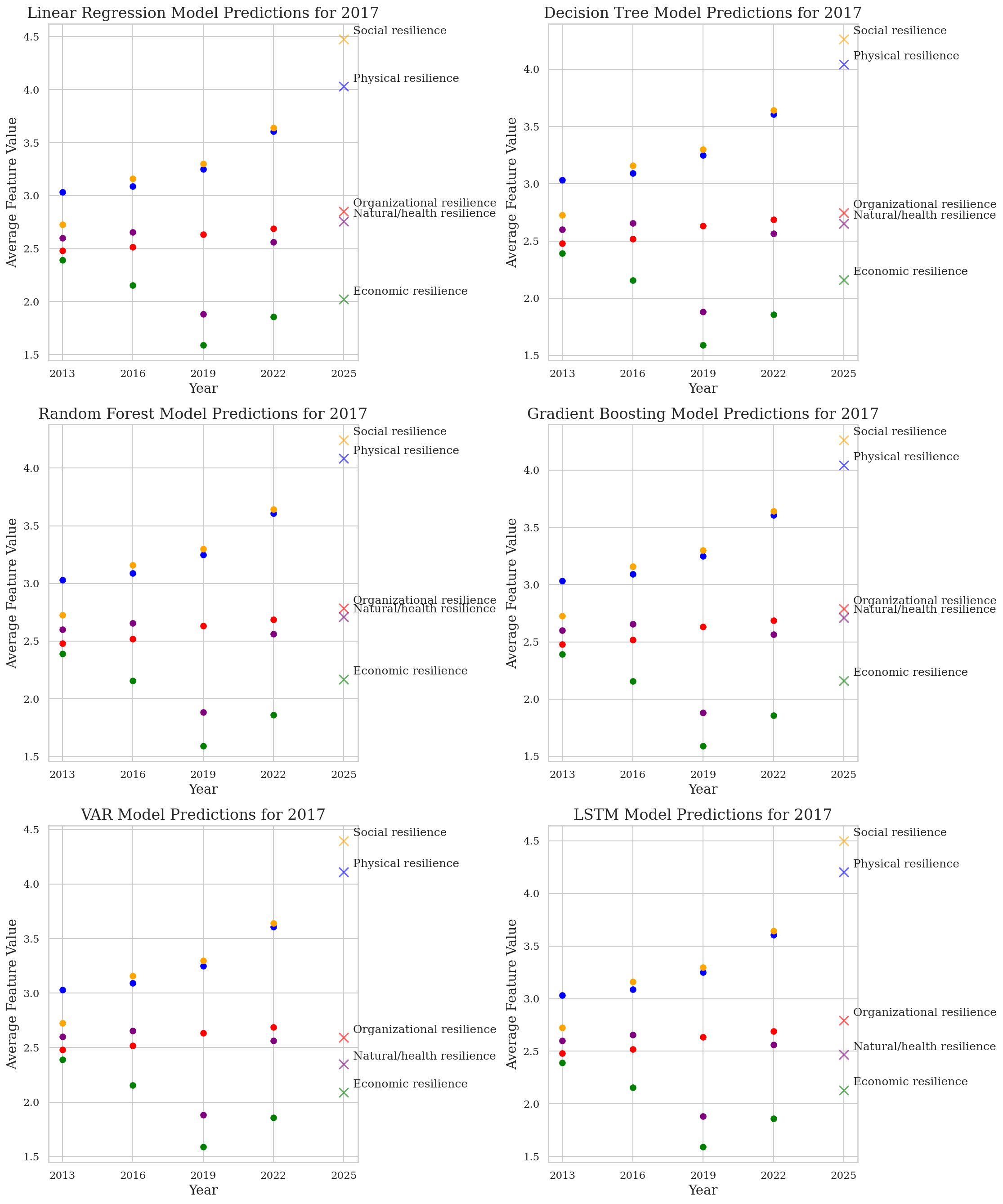}
\vspace{-5pt}
\caption{Visualizes the predicted resilience scores for each model, highlighting differences in forecasts for each resilience feature dimension.}
\label{fig:result}
\end{figure*}

Each model generated a prediction for each resilience feature in 2022. Table~\ref{tab:prediction_results} summarizes these predictions, allowing for a comparative analysis of each model's performance. Figure~\ref{fig:result} visually represents these predictions, highlighting each model's forecast for resilience features. The training process for LSTM models involves multiple epochs, as shown in Figure~\ref{fig:lstm_loss}, which illustrates the model's convergence as the loss decreases over time \cite{hochreiter1997long}.

\subsubsection{Model Performance Visualization}
To understand the training performance of the LSTM model, Figure~\ref{fig:lstm_loss} displays the loss per epoch, illustrating how the model’s accuracy improves over the training period. This figure highlights the optimization process, where the model adjusts weights to minimize prediction errors gradually. This predictive analysis demonstrates the potential of machine learning models to forecast urban resilience features accurately. Comparing the results of six different models allows for the identification of trends and the highlighting of areas that may require additional support to enhance climate resilience in urban areas. Machine learning empowers urban planners to make informed, proactive decisions, enhancing resilience planning and resource allocation.

\section{Conclusion}

This study identifies the Climate Disaster Resilience Index (CDRI) as the most compatible model for assessing urban flood resilience in Tehran’s District 6. Recognizing the limitations of the static CDRI model, the approach was enhanced by incorporating temporal data at three-year intervals from 2013 to 2022 and applying machine learning techniques to create a dynamic Temporal CDRI model that predicts total resilience in 2025. This new model captures evolving resilience conditions across different dimensions over time. The analysis indicates that in 2019, the economic and health resilience of District 6 declined, likely due to pressures from the 2019 pandemic, while other dimensions—physical, social, and organizational—showed improvement, leading to a modest overall resilience increase. By integrating machine learning, the Temporal CDRI model offers a more precise and adaptable approach for urban resilience assessment, enabling policymakers to respond proactively to changing urban conditions and better manage flood risks in Tehran’s critical regions. This study underscores the potential of AI-enhanced resilience frameworks to guide effective urban planning in flood-prone areas.

\end{document}